\documentclass[twocolumn, 5p]{elsarticle}

\usepackage{wrapfig}
\usepackage{times}
\usepackage{epsfig}
\usepackage{graphicx}
\usepackage{amsmath}
\usepackage{amssymb}
\usepackage{amsmath} % DO NOT CHANGE THIS
\usepackage{xspace}
\usepackage{comment}
\usepackage{nicefrac}
\usepackage{subcaption}
\usepackage{rotating}
\usepackage{natbib}
\usepackage{comment}

\usepackage{hyperref}

\newcommand{\mEntr}{\ensuremath{W_\sigma}\xspace}

\newcommand{\mName}{\ensuremath{W}\xspace}

\newcommand{\x}{\mathbf{x}}

\newcommand{\val}{\nu}

\newcommand{\vct}[1]{\ensuremath{\boldsymbol{#1}}}

\newcommand{\set}[1]{\ensuremath{\mathcal{#1}}}

\newcommand{\argmax}{\operatornamewithlimits{\arg\,\max}}

\newcommand{\ie}{{i.e.}\xspace}

\usepackage[draft,inline,nomargin,index]{fixme}
\fxsetup{theme=color,mode=multiuser}
\FXRegisterAuthor{tl}{Taesung}{\color{blue}Taesung}
\FXRegisterAuthor{kg}{Kathrin}{\color{cyan}Kathrin}

\begin{document}

\begin{frontmatter}

%BB: length between 5K and 10K words (check how many pages it corresponds to...)

\title{Backdoor Smoothing: Demystifying Backdoor Attacks on Deep Neural Networks
}

%\maketitle 
\address[1]{Department of Electrical and Electronic Engineering, University of Cagliari,
		Piazza d’Armi 09123, Cagliari, Italy}
\address[2]{IBM T. J. Watson Research Center, 1101 Kitchawan Road, Yorktown, New York, US}
\address[3]{CISPA Helmholtz Center for Information Security,
Stuhlsatzenhaus 5, Saarbrücken, Germany}
	\address[4]{Pluribus One, Italy}
	\cortext[cor1]{Corresponding author. Work was done while the first author was interning at IBM T. J. Watson Research Center.}
	\author[1]{Kathrin Grosse\corref{cor1}\fnref{fn1}}
	\ead{kathrin.grosse@unica.it}
	\author[2]{Taesung Lee}
	\author[1,4]{Battista~Biggio}
	\author[2]{Youngja Park}
	\author[3]{Michael Backes}
	\author[2]{Ian Molloy}

\begin{abstract}
Backdoor attacks mislead machine-learning models to output an attacker-specified class when presented a specific trigger at test time.
These attacks require poisoning the training data to compromise the learning algorithm, e.g., by injecting poisoning samples containing the trigger into the training set, along with the desired class label.
Despite the increasing number of studies on backdoor attacks and defenses, the underlying factors affecting the success of backdoor attacks, along with their impact on the learning algorithm, are not yet well understood.
In this work, we aim to shed light on this issue by unveiling that backdoor attacks induce a smoother decision function around the triggered samples -- a phenomenon which we refer to as \textit{backdoor smoothing}. 
To quantify backdoor smoothing, we define a measure that evaluates the uncertainty associated to the predictions of a classifier around the input samples. 
 Our experiments show that smoothness increases when the trigger is added to the input samples, and that  this phenomenon is more pronounced for more successful attacks.
 We also provide preliminary evidence that backdoor triggers are not the only smoothing-inducing patterns, but that also other artificial patterns can be detected by our approach, paving the way towards understanding the limitations of current defenses and designing novel ones.
\end{abstract}


\end{frontmatter}

\section{Introduction}
In the cybersecurity domain, backdoors are usually defined as covert methods to bypass authentication or encryption. This notion has been recently extended to poison machine-learning models and, in particular, deep neural networks: a semantically unrelated pattern is added to some samples of the training data, jointly with a label of a particular class. At test time, regardless of the input, the network is circumvented: it will only output the previously specified class, as long as the trigger is present~\citep{chen2017targeted,liu2017trojaning,gu2019badnets,zhu2019transferable}. 
To run such an attack, the adversary must be able to inject training samples along with the desired labels~\citep{gu2019badnets}. If the adversary cannot control the labeling process of training samples directly, a clean-label backdoor attack may be alternatively  staged~\citep{shafahi_poison_2018,zhu2019transferable}.
Other settings assume that the adversary trains the network for the user and implants the trigger into the learned weights without harming accuracy on untriggered data~\citep{liu2017trojaning}.
In other words, backdoor attacks can be staged in a plethora of diverse scenarios, including when model training is outsourced to an untrusted third party (e.g., a cloud provider), publicly-available pretrained models are used, or when training data is collected from untrusted sources.

As these settings are not uncommon nowadays, many defenses have been proposed. Some leverage the fact that the inserted trigger can be recomputed~\citep{liu2019abs,wang2019neural}, or that the internal representation of data with and without the trigger differs~\citep{chen2018detecting,tran2018spectral}.
Other works show that the backdoor might be unlearned during training or fine-tuning, if the trigger is not present~\citep{liu2018fine,li2021neural}.
However, many defenses have been shown to be not robust against attacks  slightly altered to bypass them~\citep{tan2019bypassing,nguyen2021wanet}. This arms race between defenses and adaptive attacks emphasizes how little is known about how backdoor attacks work, similarly to what has been observed for adversarial examples~\citep{biggio14-tkde,DBLP:journals/pr/BiggioR18,athalye18,gilmer18}.

The underlying intuition suggests that backdoor attacks work by introducing a strong correlation between the trigger and the attacker-specified class label, which is then picked up by the model at training time.
The fact that the trigger breaks the semantics (e.g., a dog image with the trigger mislabeled as a cat) does not affect the accuracy of the model on clean data samples, as neural networks have been shown to be extremely flexible and essentially able to learn any labelling for given data~\citep{zhang2016understanding}. 
Yet, little is known beyond this intuition.
A current hypothesis on backdoors in transfer learning points out that the amount of fine-tuning data is smaller than the available parameters, leading to overfitting~\citep{shafahi_poison_2018}.
Similar hypotheses, brought forward by \cite{wang2019neural} and \cite{zhu2021clear}  state that backdooring introduces `shortcuts' between different classes. 
One interpretation of the latter hypotheses is thus that the decision surface around backdoors is  less smooth, as the backdoor samples with the trigger live close 
to their original class. 
To summarize, despite the existence of several hypotheses, the underlying factors affecting the success of backdoor attacks, along with their impact on the learning algorithm, are not yet well understood.

In this work, we aim to overcome this limitation and shed light on how backdoors work. 
In particular, conversely to the common intuitions detailed before, we show that backdoor attacks work by inducing a smoother decision function around the triggered samples -- a phenomenon which we refer to as \textit{backdoor smoothing}. 
To characterize the phenomenon of backdoor smoothing, we first define a measure to quantify the local smoothness of the decision function (Section~\ref{sec:w}), inspired by recent work on randomized smoothing~\citep{DBLP:conf/icml/CohenRK19,lee2019tight,salman2019provably}.
Randomized smoothing has been used as a certified defense against adversarial examples~\citep{DBLP:conf/icml/CohenRK19,lee2019tight,salman2019provably}, and against label-flip poisoning attacks on linear classifiers~\citep{rosenfeld2020certified}.
To apply randomized smoothing, we perturb a classifier's input sample using Gaussian noise and observe the resulting output distribution.
We leverage this underlying idea of randomized smoothing to estimate the approximated distribution of class probabilities, and then use entropy to compute the smoothness of the classifier. 
If the output is very smooth and the classifier only outputs a single class, the measure is zero, hence quantifying the \textit{wobbliness} of the decision surface. For this reason, we name our measure \mName.\footnote{\mName is named in honor of Doctor \textbf{W}ho, who first recognized wibbly wobbly timy wimy properties.}

We then run an extensive experimental analysis and show that the decision function around benign, clean test data is less smooth than around test data containing a functional trigger (Section~\ref{sec:exp}). 
In other words, when a functional trigger is present, additional noise rarely changes  the classification output, if at all. More specifically, smoothness correlates with the accuracy of the backdoor trigger: the smoother the decision function is around the input sample,
the higher the accuracy of the backdoor.  
We further investigate whether backdoors are the only, reliable cause for backdoor smoothing, or if other \textit{smoothness-inducing patterns} may exist.
We discover that other artificially-crafted patterns may also map samples to similarly smooth regions of the classifier. For example, some perturbations applied to craft adversarial examples~\citep{biggio13-ecml,szegedy2013intriguing} tend to increase smoothness of the decision function around the input data, as well as backdoor triggers that are either verified or reconstructed by defensive mechanisms~\citep{gao2019strip,liu2019abs,wang2019neural,zhu2020gangsweep,xiang2020detection,xiang2021reverse}. Intriguingly, even when no backdoor attack is staged during training, some defenses may reconstruct effective triggers, \ie, adversarial perturbations that mislead classification at test time. Our approach confirms that they induce smoothness in the decision surface (the related implications are discussed in Section~\ref{sec:implications}).
To summarize, we believe that our measure $\mName$ could possibly help detect anomalous behaviors beyond those exhibited by backdoor attacks, including more generic \textit{smoothness-inducing patterns} that are able to subvert predictions, thus fostering the development of more effective and general defensive strategies.

We conclude this work by discussing related works (Section~\ref{sec:reWork}), along with our main contributions and promising future research directions (Section~\ref{sec:conclusions}).

\section{Backdoor Smoothing and Wobbliness} %1 page
\label{sec:w}

We introduce a measure to quantify the local smoothness of the decision function of a classifier, in other words a measure connected to the phenomenon of \textit{backdoor smoothing}. We first provide a high-level intuition of how the measure works, along with the necessary notation. Afterwards, we motivate and formalize our \textit{wobbliness} measure \mName. 

\begin{figure}
\centering
\includegraphics[width=1.0\linewidth]{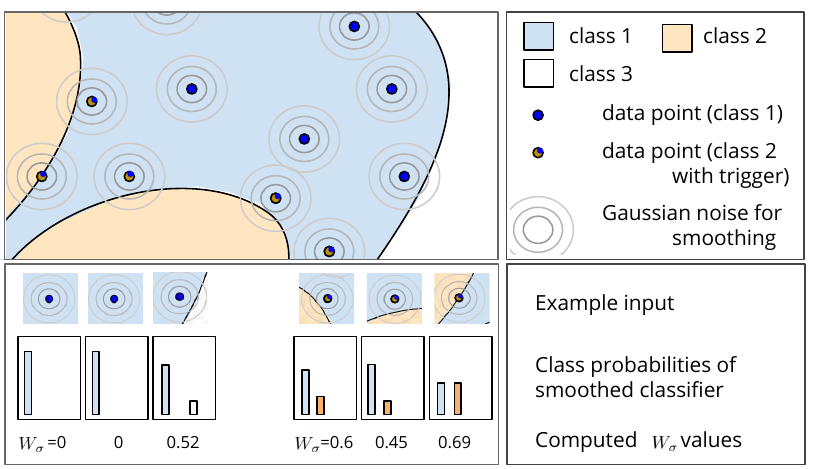}
    \caption{
    Differing local smoothness around points with and without trigger (top)  and the computation of \mEntr (bottom) to distinguish them. The setting is inspired by the hypothesis that backdoored points (yellow with blue) live closer to their original class (yellow area) than benign test data (blue). Hence, the area around backdoors is not smooth. Benign test data instead live in a smooth environment.
    Our measure allows to distinguish these two cases.
    To compute our measure, we sample Gaussian noise around the test point (gray circles), and compute the class probabilities of the resulting approximated smoothed classifier (visualized as histograms). We then compute the entropy of the class probabilities and depict them below the probability distribution. \mEntr, or the resulting values reflect the difference in smoothness between the two cases.}  
    \label{fig:WWTW_intuition}
\end{figure}

\paragraph{Intuition} Our goal is to measure the smoothness around data with and without an active trigger. 
 We present an intuition in Figure~\ref{fig:WWTW_intuition}, where backdoored data (yellow dots with blue) live close their original class (yellow) in a non-smooth area. Clean test data (blue dots), instead, live in a smooth area. 
 Our measure approximates a smooth classifier by sampling Gaussian noise around the test point (gray circles). We then compute the distribution over the output classes and derive a measure that describes the smoothness of this output distribution, as visible at the bottom of Figure~\ref{fig:WWTW_intuition}. 
 If for example only one class is outputted across the sampled ball, the measure is zero. 
 Our measure, \mName, when computed on a batch of points, yields insights about the smoothness of the area the input data lives in.

\paragraph{Notation} Let us assume that we are given an input sample $\vct x \in \set X \subseteq \mathbb R^d$ of $d$ dimensions to be classified as one of the $c$ classes $\{1, \ldots, c\}$. 
The classification function is represented as $g_k : \set X \mapsto \mathbb R$, providing a continuous-valued confidence score for each class $k=1, \ldots, c$. The classifier will assign the input sample $\vct x$ to the class $k$ exhibiting the highest classification confidence, \ie, $k = \argmax_{j=1, \ldots, c} g_j(\vct x)$. We can thus define $f_k$ as the one-hot encoding of the predicted class labels as:
\begin{equation}
f_k(\vct x) =
\left\{
	\begin{array}{ll}
		1 \, ,  & \mbox{if } k = \argmax_{j=1, \ldots, c} g_j(\vct x) \, , \\
		0 \, ,& \mbox{otherwise \,.} 
	\end{array}
\right. 
\end{equation}

\paragraph{Smoothing} Let $\vct \delta \sim \set N(\vct 0, \sigma^2 \mathbb I) \in \mathbb R^d$, \ie, an isotropic Gaussian perturbation parameterized by its standard deviation $\sigma$. We denote the probability of classifying $\vct x + \vct \delta$ as class $k$ with
\begin{equation}
    p_k(\vct x) = \mathbb E_{\vct \delta \sim \set N(\vct 0, \sigma^2 \mathbb I)} f_k(\vct x+\vct \delta) \simeq \frac{1}{n} \sum_{i=1}^n f_k(\vct x + \vct \delta_i) \, ,
\end{equation}
where each probability value $p_k(\vct x)$ depends on the choice of $\sigma$ and the amount of sampled points $n$. Intuitively, the larger $n$, the more accurate are the approximated class probabilities $p_k(\vct x)$. To understand the effect of $\sigma$,
consider the third clean test point in Figure~\ref{fig:WWTW_intuition}. If $\sigma$ is very small, the sampled ball is classified entirely as blue class. As $\sigma$ increases, the white class' probability will be nonzero, too. 

We are interested in the smoothness of the classification probabilities. As the entropy is zero if only one class has all the probability mass, and is maximal when $p_1(\vct x) = p_2(\vct x) = \dots = p_c(\vct x)$, we define our measure $W$ as the entropy of the smoothed probability outputs,
\begin{equation}
    W_\sigma(\vct x) = - \sum_{k=1}^c p_k(\vct x) \log{p_k(\vct x)} \, .
\end{equation}
We further write $\mEntr$ to highlight the dependency of $W$ on $\sigma$, using the latter as subscript. To conclude, we would like to emphasize that as the measure is zero at the maximal smoothness, and maximal when the surface is very non-smooth, it actually quantifies the \textit{wobbliness} of the decision surface. In reference to this wobbliness, we name the measure \mName. 

\mName is thus well suited to study backdoor smoothing: if the measure is low for a set of points, we can deduce that the local decision surface is very smooth around these points. We now turn to our empirical results when analysing backdoored neural networks
using \mEntr.

\section{Empirical Evaluation}
\label{sec:exp}
We first evaluate in Section~\ref{sec:param} \mEntr's parameters and their influence on the measurement. Afterwards, in Section~\ref{sec:backdoorsSetting}, we study backdoors though the lens of \mEntr.

\subsection{Parameters of \mName} \label{sec:param}
Before investigating backdoor attacks using \mEntr,  we empirically validate our understanding of its parameters to motivate our choices of these parameters later on. As stated in Section~\ref{sec:w}, \mEntr relies on two important parameters. 
This includes the variance of the noise distribution $\sigma$ which influences how local the smoothed classifier is. Further, the amount of sampled noise points $n$ per test point influences \mEntr. We show that as $n$ is larger, the measure is more stable. Yet, an $n$ as small as $250$ points suffices to obtain a stable output of the measure. 
We first outline
the experimental setting and the layout of the plots, before we turn to the parameters $\sigma$ and $n$. Afterwards, we proceed with our study on backdoors. 

\paragraph{Experimental Setting}
We study two datasets, Fashion MNIST~\citep{xiao2017/online} and CIFAR10~\citep{krizhevsky2009learning}. On the former,
we deploy small networks which achieve an accuracy of around $88$\%.
These networks contain a convolution layer with $32$ $3\!\times\!3$ filters, a max-pooling layer of $2\!\times\!2$, another convolution layer with $12$ $3\!\times\!3$ filter,
a dense layer with $50$ neurons, and a softmax layer with  $10$ neurons.
We further experiment on CIFAR10, where we 
train a ResNet18~\citep{he2016deep} $200$ epochs to achieve an accuracy of $91.8$\%. 
In general, we report the results of one network in this subsection, yet rerun the experiments several times to confirm that the results remain consistent. 

\paragraph{Plots} Given a batch $X$ of size 250 test points, we plot the distribution of $\mEntr(X)$ using box plots. These plots depict the mean (orange line), the quartiles (blue boxes, whiskers) and outliers (dots).
We follow a standard definition for outliers by \cite{frigge1989some}: An outlier is defined as a point further away than 1.5 the interquartile range from the quartiles.
More concretely, $Q_{25}$ is the first quartile and $Q_{75}$ is the third quartile (and $Q_{50}$ is the median). Value $\val$ is an outlier if and only if
\begin{equation}
\val > Q_{75} + 1.5 \times (Q_{75}-Q_{25}) \text{\space or \space} \val < Q_{25} - 1.5 \times (Q_{75}-Q_{25}) \text{\space ,}
\end{equation}
in other words if
$\val$ is more than $1.5$ times the interquartile range $(Q_{25}-Q_{75})$ away from either quartile $Q_{25}$ or  $Q_{75}$. 

\begin{figure}
\centering
    \includegraphics[width=0.49\linewidth ]{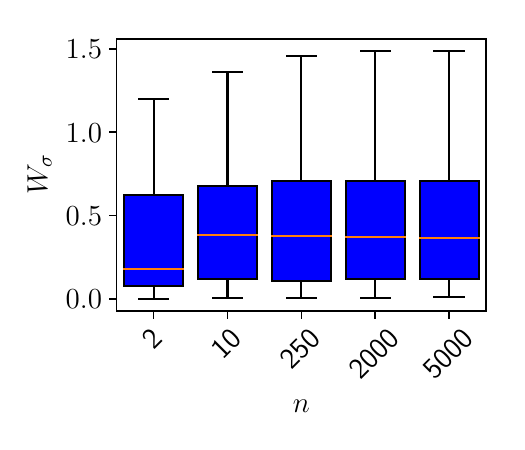}
    \includegraphics[width=0.49\linewidth ]{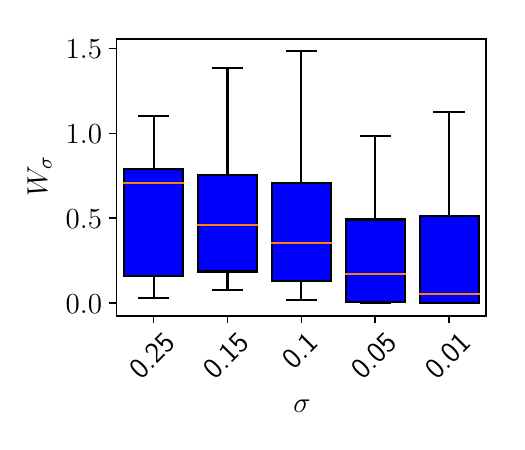}
    \includegraphics[width=0.49\linewidth ]{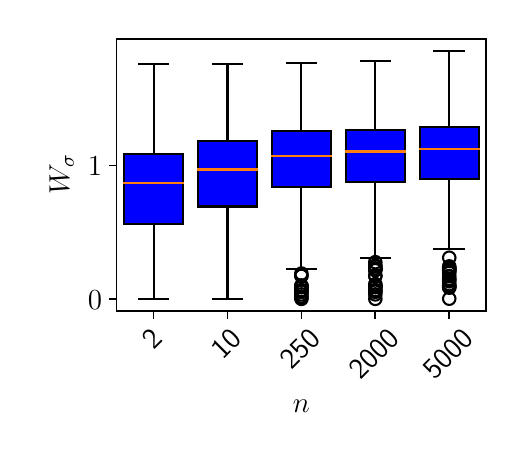}
        \includegraphics[width=0.49\linewidth] {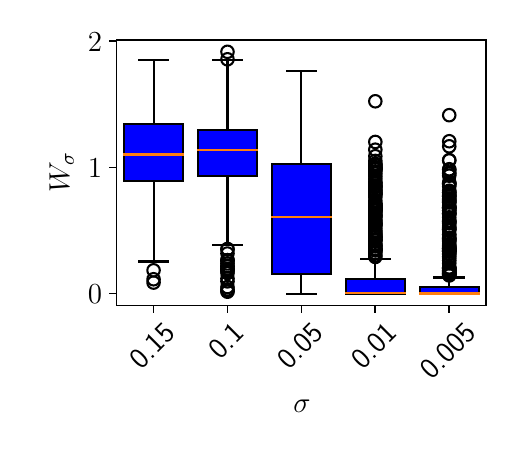}
    \caption{Box-plots reporting \mEntr values on clean samples from Fashion MNIST (\textit{top row}) and CIFAR10 (\textit{bottom row}), as a function of the number of noised points $n$ (\textit{left}) and $\sigma$ (\textit{right}).
    } 
    \label{fig:Msigma}
\end{figure}

\paragraph{Parameters $\mathbf{\sigma}$ and $\mathbf{n}$}
Both parameters are visualized in Figure~\ref{fig:Msigma}.
We first describe the effect of $n$, the number of sampled points around each given test point. Independent of the dataset used, as $n$ increases, the mean of the computed \mEntr values increases and the spread decreases. Starting from roughly 250 sampled points, the results are relatively stable, and the changes in mean and spread are rather small.  We conclude that a rather small number of sampled points around each test point, e.g. $250$, is sufficient to approximate the classifier for our measure.

We verify our intuition about $\sigma$. As our features are in range $[0,1]$, we also choose $\sigma \in [0,1]$, where we do not truncate the noise points if they are outside $[0,1]$ after adding the noise. As CIFAR10 has a larger features space due to the images being colored, we plot the results for smaller values of $\sigma$ in this case. 
Independent of the dataset, as the $\sigma$ decreases, the spread of the mean of \mEntr values decreases, until there is no more variance at $\sigma=0.01$. This is expected, as we sample a smaller and smaller radius. Yet, the transition is smoother on Fashion MNIST than on CIFAR10. On the latter, the results for $\sigma = 0.15$, $\sigma = 0.1$, $\sigma = 0.01$ and $\sigma = 0.005$ are similar.  

\paragraph{Conclusion on Parameter Choice} When re-sampling each point in batch $X$ at least $250$ times, $\mEntr(X)$ is sufficiently stable. Furthermore, $\sigma$ as a parameter allows us to measure the uncertainty of the smoothed classifier at a different granularity.

\subsection{Backdoor Smoothing}\label{sec:backdoorsSetting}
We are now equipped with an understanding of \mEntr that allows us to study backdoor smoothing. In Section~\ref{sec:expSetting}, we first describe the experimental setting of the following experiments. Afterwards, in Section~\ref{sec::initialWExp}, we show that backdoor smoothing, or \mEntr changes between a backdoor and clean test data. In Section~\ref{sec:backdoorAccuracyMentr}, we deepen this understanding and correlate backdoor success with \mEntr and backdoor smoothing. Finally, in Section~\ref{sec:smoothingNecCondition}, we investigate whether backdoors are a necessary condition for backdoor smoothing. In other words, we first verify that backdoors cause backdoor smoothing and then test whether other patterns induce backdoor smoothing as well. 

\subsubsection{Experimental Setting}\label{sec:expSetting} However, we first detail the threat model and experimental setup before we describe the results of our qualitative study. 

\begin{figure}
    \includegraphics[width=\linewidth]{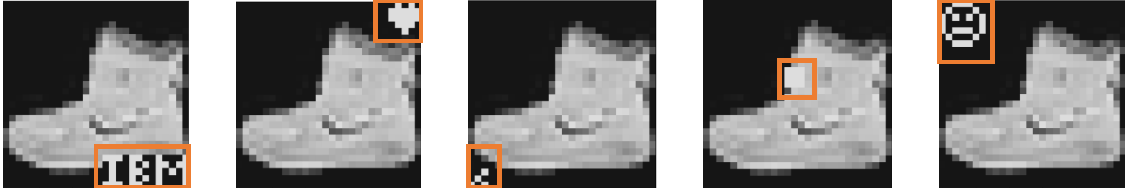}
    \caption{Possible backdoors on Fashion MNIST.} 
    \label{fig:vackdoorEx}
\end{figure}

\paragraph{Attacker and Threat Model} 
A range of threat models have been studied in the literature, depending on whether the victim uses a pre-trained model or trains the model himself.
Furthermore, the attacker can target one or several classes with the inserted backdoor.
In other words, both the percentages of backdoor points in training as well as the number of targeted classes may vary.
To avoid that any of these factors influences \emph{backdoor smoothing}, we capture each of the previously mentioned factors in the following three threat models:
%In this work, we study the following three threat models for planting backdoors, respectively denoted with T1, T2 and T3.
\begin{enumerate}
    \item[T1.] The victim is unable to inspect the training data, as they receives only the pre-trained model. The attacker is thus able to poison a large fraction of the training data and targets one class, as done in the work by \cite{wang2019neural} and \cite{chen2018detecting}. We poison 15\% of the training data, choose one target class and add the trigger to all classes except the target class.
    \item[T2.] This is a variation of the first threat model and was introduced by \cite{gu2019badnets}. As before, we poison
    15\% of the training data, but the backdoor is added to each class. The output is now determined as the previous class $c+1$ modulo the number of classes. 
    \item[T3.] The attacker alters the training data, because the victim trains the model themselve. 
    As the victim might inspect the data before training the model, the amount of injected poisons is small (roughly 250 for more than 50,000 training points) as done by \cite{tran2018spectral} and \cite{chen2018detecting}.
\end{enumerate}
 
In all cases, a backdoor pattern is added to a portion of the training samples. Examples of such a pattern on image classification tasks are shown in Figure~\ref{fig:vackdoorEx}. Many of these backdoors are also used by \cite{chen2018detecting} and \cite{gu2019badnets}. We draw a backdoor pattern from this set, and overwrite the original pixels in the lower right corner with the corresponding trigger.

\paragraph{Model and Datasets}
We train a small convolutional neural network on Fashion MNIST that is able to learn the backdoors well: a convolution layer ($64$ $3\!\times\!3$ filters), a max-pooling layer ($2\!\times\!2$), another convolution layer ($32$ $3\!\times\!3$ filters), two dense layers with $500$ neurons are followed by a softmax layer with $10$ neurons.
To reduce the impact of randomness, we train and repeat the experiments multiple times.
The trained models achieve around $90$\% accuracy on benign data, and $99$\% on inputs with the backdoors.
On the CIFAR10 data, we use a ResNet18 architecture. 
As the smoothness of the classifier might be affected by overfitting, and we assume the small classifier above to overfit,  we train the CIFAR networks few epochs to underfit the benign data at around $60$\% accuracy. The CIFAR networks however perform in general well, or with accuracy of $>99$\%, on test points with an active trigger.  

\paragraph{\mEntr Parameters}
As we have seen in Section~\ref{sec:param}, \mEntr is sufficiently stable we re-sample each input point $250$ time or more often. We hence set $n=250$ unless otherwise mentioned. More challenging is the choice of a suitable $\sigma$, which we will tackle in our first experiment in the next subsection.

\begin{figure}
\centering
\begin{subfigure}{0.48\linewidth}
    \includegraphics[width=\linewidth]{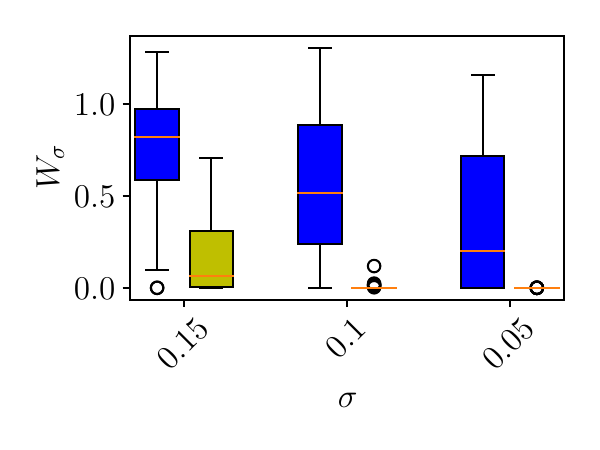}
    \caption{Functional trigger}
\end{subfigure}
\begin{subfigure}{0.48\linewidth}
    \includegraphics[width=\linewidth]{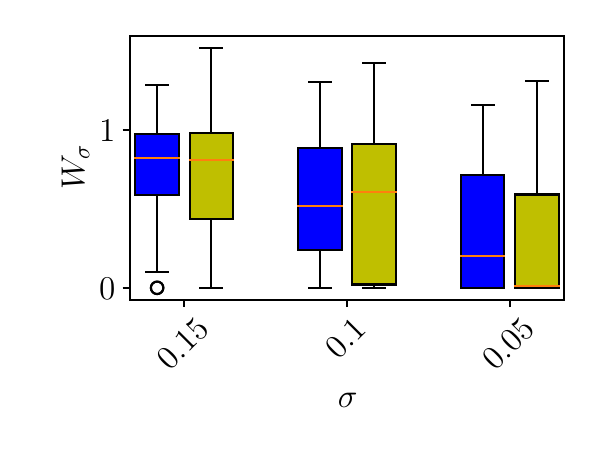}
        \caption{Benign data}
\end{subfigure}
    \caption{\mEntr and backdoors on Fashion MNIST. We compare clean test data (blue) and a functional trigger (99\% accuracy, left plot)/an unused trigger (9\% accuracy on poisoning labels, 89\% on clean labels, right plot).}   \label{fig:backdoorWWTW}
\end{figure}

\subsubsection{Effect  of Backdoors on the Decision Surface}\label{sec::initialWExp}
Before we deepen our understanding on  the correlation between \mEntr and backdoor accuracy and which samples induce backdoor smoothing, we first need to understand at which locality differences arise.
We thus first show the characteristics of samples around a backdoor and how \mEntr can capture them at different localities, depending on the parameter $\sigma$.

\paragraph{Experimental Setting} We consider two batches with 25 randomly chosen test points. The first batch $X_C$ contains clean test data, the second, $X_A$, contains test data with an active trigger the network was trained on.
The trigger is the leftmost pattern from Figure~\ref{fig:vackdoorEx}, and used in training as specified in threat model T1.
We plot the distributions of $\mEntr(X_C)$ (blue) and $\mEntr(X_A)$ (yellow) using the box-plots from the previous experiments in
Figure~\ref{fig:backdoorWWTW} on the left. 
As a sanity check, we repeat the experiments using an additional batch $X_{NA}$ where we add the fourth pattern from Figure~\ref{fig:vackdoorEx} as an inactive trigger. In other words, the network has not been trained on this trigger pattern. We draw a new batch $X_C$ and plot $\mEntr(X_C)$ (blue) and $\mEntr(X_{NA})$ (yellow) on the right in Figure~\ref{fig:backdoorWWTW}.

\paragraph{Results} We find that the local uncertainty of the smoothed classifier, measured by \mEntr, is largely different for data with and without the active trigger pattern.
On data with no trigger or an inactive trigger, \mEntr indicates that the decision surface is not very smooth. 
On the other hand, and in particular at small $\sigma$, the distribution of \mEntr shows that the smoothed classifier is very certain on the data with the active trigger.

\paragraph{Conclusion} Using our measure $\mEntr$, we found that the classification output remains consistent regardless of the noise added to compute $\mEntr$, and the decision surface around the sample with an active trigger is very smooth.
This behavior is the attacker's goal: as soon as the
backdoor is present, other features become irrelevant.
We call this property \emph{backdoor smoothing}. 

\begin{figure}
\centering
\scriptsize{Kendall's $\tau$: $-0.5$, $p=0.001$ 
\hspace{4.3em} Kendall's $\tau$: $-0.49$, $p=1.5e^{-10}$\\}
\scriptsize{Pearson corr.: $-0.77$, $p=7.5e^{-5}$ 
\hspace{3.1em} Pearson corr.:  $-0.58$, $p=1.8e^{-8}$}

\includegraphics[width=0.49\linewidth]{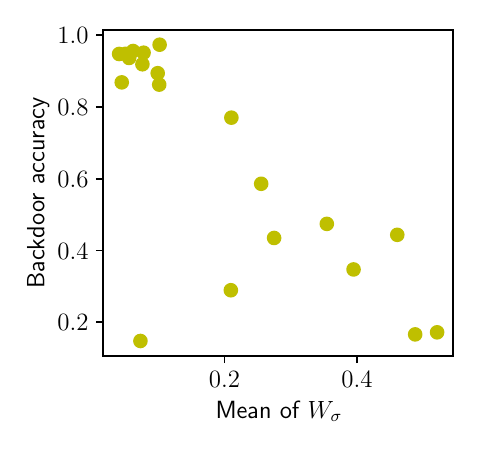}
\includegraphics[width=0.49\linewidth]{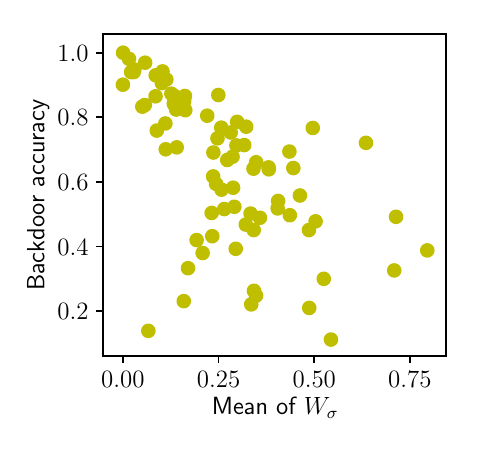}
\caption{Backdoor accuracy and average \mEntr on CIFAR10 for threat model T1 (\textit{left}) and  Fashion MMIST for T1 (\textit{right}).
 Above the plots, we write the computed p-values for Kendall's $\tau$ and Pearson correlation, two tests with H$_0$ that no correlation is present.}   \label{fig:backdooracctradeOff}
\end{figure}

\subsubsection{Backdoor Accuracy and \mEntr}\label{sec:backdoorAccuracyMentr}

In the previous experiments, we provided first evidence of \emph{backdoor smoothing}, highlighting that the decision surface is smoother around backdoored samples. 
As hypothesized, this implies that the classifier ignores the noise added to compute \mEntr. One interpretation is that this is a necessary condition, \ie, a low value of \mEntr implies that a trigger was learned well by the network. We now further validate this hypothesis.

\paragraph{Experimental Setting T1} To this end, we study T1 and train Resnets on CIFAR10.
This time, we train 12 networks on the third backdoor trigger and 12 on the fifth backdoor trigger shown in  Figure~\ref{fig:vackdoorEx}, using threat model T1. 
As before, we exclude backdoors with accuracy below random guess. As before, the smaller trigger seems harder to learn, with 4 networks not achieving random guess accuracy on the trigger. For the second trigger, all networks perform very well on the trigger.

\paragraph{Results T1} We plot the mean of \mEntr over a batch of triggered test points and the backdoor accuracy for each backdoor in the left plot in Figure~\ref{fig:backdooracctradeOff}.
As before, backdoor accuracy and the mean of \mEntr seem negatively correlated. If the backdoor accuracy (y-axis) is very high, then the mean of \mEntr (x-axis) is low (left upper corner in the plot). Few points exhibit a backdoor accuracy below 0.4, these are spread far apart. As before, we validate correlation using Kendall's $\tau$ and Pearson correlation. 
The correlation is quantified with -0.5 (Kendall's $\tau$) and -0.77 (Pearson), and again negative.
The p-values are $0.001$ (Kendall's $\tau$) and $7.5e^{-5}$ (Pearson correlation), and thus confirm that backdoor accuracy and the mean of \mEntr are negatively correlated. Both p-values are larger than in the previous case, which is most likely due to the smaller sample size (17 cases here vs 79 cases in the experiments on Fashion MNIST).

\paragraph{Experimental Setting T2} We repeat the experiments with 5 networks on the third backdoor trigger, and 5 networks on the fifth backdoor trigger from Figure~\ref{fig:vackdoorEx} on Fashion MNIST, using threat model T2.
As we target each class, 
we take a batch of test points with active trigger for each class $c$, $X_c$, and compute $\mu (\mEntr(X_c))$, e.g. the mean of \mEntr on the batch. We compute Backdoor accuracy separately for each class and exclude backdoors with an accuracy below random guess. For the small trigger, these are 19 out of 50 backdoor/target class pairs. For the larger smiley trigger, 2 are excluded. This reflects that the smaller trigger, as it has fewer features, is harder to learn for the networks. 

\paragraph{Results T2} We visualize the results in the right plot in Figure~\ref{fig:backdooracctradeOff}.
The accuracy and the mean of \mEntr seem negatively correlated. If the backdoor accuracy (y-axis) is very high, then the mean of \mEntr (x-axis) is low (left upper corner in the plot). As the mean of \mEntr values increases, the accuracy decreases and also varies more. To validate our intuition that both quantities are correlated,  we compute  Kendall's $\tau$~\citep{kendall1938new} and Pearson correlation~\citep{pearson1895notes} on our sample. Both tests assume no correlation between the two features (backdoor accuracy and \mEntr) as H$_0$. Both tests reject the H$_0$ with very small p-values of $1.5e^{-10}$ (Kendall's $\tau$) and $1.8e^{-8}$ (Pearson correlation), confirming the correlation visible in the plots. The correlation is quantified with -0.49 (Kendall's $\tau$) and -0.58 (Pearson), and is indeed negative.

\paragraph{Conclusion} We conclude that the mean of \mEntr over a batch of test points with an active trigger correlates negatively with backdoor accuracy. In other words, high local smoothness correlates positively with backdoor accuracy.
This confirms our previous intuition that non-trigger features become more irrelevant when the accuracy on the trigger is higher. 
In other words,
the results confirm that \emph{backdoor smoothing} holds in general and is related to backdoor accuracy.

 \begin{figure}
 \centering
\begin{subfigure}{0.499\linewidth}
    \includegraphics[width=1\linewidth ,trim={0cm .4cm 10cm 0cm},clip]{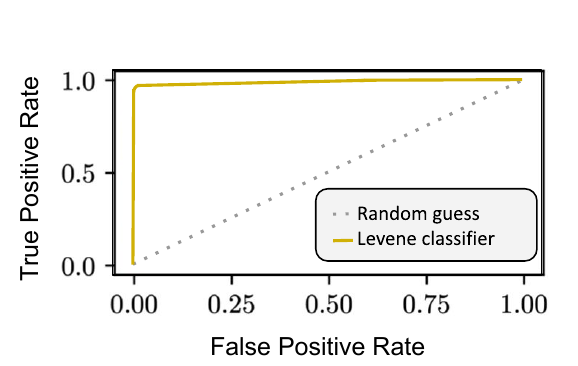}
   \caption{CIFAR10, 15\% poisons}
   \end{subfigure}
   \hspace{-0.5em}
   \begin{subfigure}{0.499\linewidth}
	\includegraphics[width=1\linewidth ,trim={10cm .4cm 0cm 0cm},clip]{figs/backdoors.pdf}
	\caption{Fashion MNIST, 250 poisons}
	\end{subfigure}
    \caption{ROC curve of a classifier based on \mEntr and the Levene test to detect differences between samples without and with active trigger (yellow) and corresponding random guess baseline (gray dots).}
    \label{fig:backdoorDetect}
    \vspace{-15pt}
\end{figure}

\subsubsection{Smoothness-inducing Patterns}\label{sec:smoothingNecCondition}
Given the differences in smoothness for benign test data and data with a trigger, the question remains whether this smoothness is unique and only caused by learned backdoors. In other words, we investigate whether backdoors are a necessary condition for backdoor smoothing.
To this end, we test on a large scale whether \mEntr differs statistically when computed on batches of test data with and without functioning trigger. To assess that this difference is consistent (e.g., \mEntr is overall smaller when the trigger in the input batch is active), we build a classifier which we evaluate based on ground truth labels. 
To further confirm that backdoors are the only patterns to introduce backdoor smoothing, we also evaluate the test on universal adversarial examples~\citep{moosavi2017universal} and unseen, crafted backdoor candidates~\citep{wang2019neural}. 

\paragraph{Experimental Setting}
We train 3 networks on clean data and 9 networks with different implanted backdoors. 
Of the latter backdoored networks, three are trained with backdoors one, three and four each from Figure~\ref{fig:vackdoorEx}.
We study two settings: T1 on CIFAR10 using ResNets and T3 on Fashion MNIST.
In each setting, \mEntr is computed on a batch of clean data for each network and on the active trigger for backdoored networks. 
To test for false positives, we also evaluate on backdoors not used in training. To this end, we also add all other five backdoors in Figure~\ref{fig:vackdoorEx}.
We feed the output of \mEntr on the corresponding batches in a statistical test. 
As opposed to a single accuracy value, we opt to plot the discriminating power of the test as a receiver operating characteristic (ROC) curve. 
This ROC curve assesses in a more general manner how well the test separates the two classes, in our case backdoored and unbackdoored networks.
More specifically, we use the ground truth knowledge from generating the data and pair active/inactive triggers with clean samples or clean with clean samples to evaluate the test.

\paragraph{Choice of Statistical Test} Given the results from Section~\ref{sec::initialWExp}, we decide to use a statistical test with the $H_0$ hypothesis that both populations have equal variance as our classifier.
We decide to use the Levene test~\citep{olkin1960contributions}
as this test can be used with a small sample size to prevent detection of overly small differences in the variances~\citep{uttley2019power}.
Although the Levene test assumes a normal distribution, which is not the case here, the test is robust if the actual distribution deviates.
The test is however sensitive to outliers, which we consequently remove. To this end, we use the definition from above by \cite{frigge1989some}. If all points have the same value, we do not remove any outliers.

\paragraph{Active Triggers and \mEntr} We plot the ROC curves in Figure~\ref{fig:backdoorDetect}.
For T1 on CIFAR10 the obtained area under ROC curve is $0.99$. In the case of T3 on Fashion MNIST the area under the ROC curve is $1.0$. Both values correspond to a perfect classifier between active backdoor triggers and untrained triggers. We thus conclude that, jointly with the observations from Section~\ref{sec::initialWExp}, backdoors indeed induce a smoother decision surface. We will now show, however, that also other patterns differ in terms of smoothness as measured by $\mEntr$.

\paragraph{Universal Perturbations and \mEntr} 
Universal adversarial examples are perturbations that are crafted after training and lead
to misclassification when added to several data points~\citep{moosavi2017universal}. 
They are thus similar in that they lead to misclassification of several data points, yet differ in that the network has not been trained on these patterns.
We repeat the previous experimental setting, however using universal perturbations instead of backdoor triggers added to the data when computing \mEntr. We then compute the Levene test, and compute the area under the ROC curve. Assuming that universal perturbations induce backdoor smoothing, we consider the two ground truth labels \textit{clean} and \textit{universal perturbation}. 

\begin{figure}
 \centering
\begin{subfigure}{0.495\linewidth}
    \includegraphics[width=1\linewidth ,trim={10cm .4cm 0cm 0cm},clip]{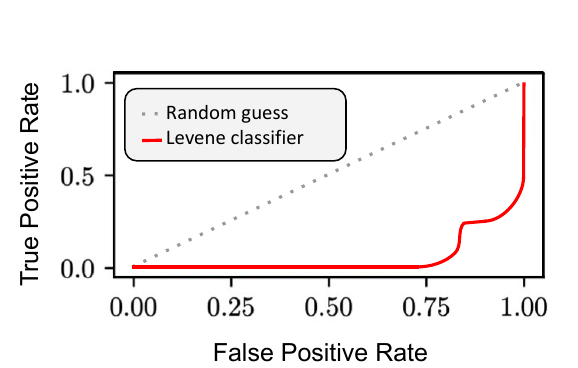}
   \caption{CIFAR10, 15\% poisons}
   \end{subfigure}
   %\vspace{-2em}
   \begin{subfigure}{0.495\linewidth}
	\includegraphics[width=1\linewidth ,trim={0cm .4cm 10cm 0cm},clip]{figs/universal.pdf}
	\caption{Fashion MNIST, 250 poisons}
	\end{subfigure}
    \caption{ROC curve of a classifier based on \mEntr and the Levene test to detect differences between samples without and with universal perturbations (red) compared to a random guess baseline (gray dots).}
    \label{fig:universalDetect}
    \vspace{-15pt}
\end{figure}

We plot the ROC curves in Figure~\ref{fig:universalDetect}. 
As visible in the left sub-figure, the area under the ROC curve for T1 on CIFAR10 is 0.05, where we make the observation that the labels are flipped: whereas active backdoor triggers showed less variances than benign data, benign data shows less variance then universal perturbations on CIFAR10. Using T3 on Fashion MNIST, however, we obtain an area under the ROC curve of $0.56$, which is comparable to random guess. In other words, the test is not able to distinguish samples with universal adversarial examples from benign data based on \mEntr. 
In other words, our results on universal perturbations are inconclusive. However, universal examples also differ from backdoor triggers in so far as they do not require all samples they are added to to be misclassified as the same class.

\paragraph{Perturbations from Neural Cleanse}
Yet, there are approach to craft a perturbation at test time with the constraint of outputting the same class for all samples. On example is Neural Cleanse~\citet{wang2019neural}, an approach to determine whether a network is backdoored by generating trigger candidates. 
The idea behind their algorithm is that a backdoor leads to a perturbation of minimal size that, when applied to any image (in a given input set), will cause \emph{all} images to return the same fixed class.
We generate a trigger candidate for each class on clean and backdoored models from our pool described above.
As before, we add the candidates to
clean test data, compute the $\mEntr$, and feed the measure into the Levene test.  
The test detects all of the inputs with the crafted perturbation as backdoors with the same, very low p-value.
As a sanity check, we add the patterns found to the training data (which has not been used to generate them) and find that the accuracy on the targeted class is 100\% in all cases. 
The crafting mechanism was thus able to find a smoothness-inducing, although the patterns were not seen during training.
We can thus refute that backdoors are a necessary condition for backdoor smoothing.

\subsection{Summary of the Results and Discussion}
In this section, we have investigated the local smoothness of networks trained with a backdoor trigger. At test time, our measure \mEntr indicates that the surface is smoother around a data point with the trigger than compared to a clean test point. We name this property \emph{backdoor smoothing}. In particular, our experiments indicate that the smoother the local decision surface around the trigger, the higher is the accuracy of the trigger on the attacker's chosen class. 
We further investigated whether an active backdoor trigger is a necessary condition for backdoor smoothing, and find this conjecture to be false. In other words, there exist other, smoothness-inducing patterns crafted after training that exhibit \emph{backdoor smoothing}. These patterns induce smoothness although the network has never been exposed to these specific patterns during training. Our findings thus indicate that \emph{backdoor smoothing} can be used to detect backdoors and also other, smoothing inducing adversarial perturbations.
This entails patterns that are generated in the context of backdoor detection, raising the question of the implications of our work for backdoor mitigations. We will discuss these implications in the next section.   
\section{Implications for the Design of Mitigations}\label{sec:implications}
Backdoor smoothing has implications for the design of mitigations against backdoor attacks if they rely for example on smoothness assumptions to generate or verify a trigger. 

\paragraph{Backdoor Smoothing and Trigger Generation} Some works explicitly use backdoor smoothing or analogous approaches to generate candidate triggers. For example \cite{wang2019neural} reverse engineer the trigger by
 collecting a batch of clean points and crafting a pattern that misclassifies all of them. Other approaches, like for example \cite{xiang2020detection} or \cite{xiang2021reverse}, craft a perturbation for each class pair, assuming the perturbation for the source and target backdoor class will be small if a trigger maps between the classes. 
 Further \cite{liu2019abs} take advantage of a similar idea, yet assuming that individual neurons provide a stable output for a backdoor.
 We have shown that patters generated by these approaches might be in fact adversarial examples~\citep{Dalvi:2004:AC:1014052.1014066,biggio13-ecml,szegedy2013intriguing} that were not implanted into the training data. In this sense, the difference between a backdoor and  another, smoothness-inducing pattern or adversarial example exhibiting \emph{backdoor smoothing} is not necessarily clear.  

\paragraph{Backdoor Smoothing and Trigger Verification} 
Analogous implications hold for approaches relying on smoothness to decide whether a pattern is a trigger or not.
For example, \cite{gao2019strip} propose to perturb the input using superimposition of two images and then evaluate the consistency of the output classes. If one class prevails (e.g., the surface is smooth), the presence of a trigger is deduced. Analogously, \cite{zhu2020gangsweep} compute the variance of the logits between a clean image and an image with a universal adversarial perturbation~\citep{moosavi2017universal} added.
Here as well, the pattern is confirmed to be smoothness-inducing, and might be a backdoor or an adversarial example. In other words, there is no guarantee the found pattern was inserted with malicious intention into the training data.  
 
 \paragraph{Conclusion} When no ground truth information on a trigger pattern is available, the question whether a smoothness-inducing pattern was indeed implanted in the training data cannot be reliably answered. 
Some approaches tackle this issue by making additional assumptions on the trigger size~\citep{wang2019neural} or incorporate more information like explanations and smoothness of the trigger pattern~\citep{guo2019tabor}.

To finish the section, we would like to remark that there are conceptually unrelated defenses. To name a few examples, \cite{zhao2020bridging} compute a new model by interpolating in the loss space between two (potentially) backdoored models. Furthermore, defenses relying on pruning or retraining~\citep{liu2018fine,li2021neural, aiken2021neural} are not related. These defenses have to be thoroughly evaluated depending on their own specifics. 
\section{Related Work}\label{sec:reWork}
We first review related work in the area of backdoors. Afterwards, we describe randomized smoothing, a conceptually related technique to obtain robustness against perturbation at test time. To conclude, present input sensitivity measures, as some are also related to the smoothness of the classifier.

\paragraph{Backdoor Attacks} After the initial work on backdoor attacks~\citep{gu2019badnets,chen2017targeted}, many defenses were proposed~\citep{wang2019neural,liu2019abs,tran2018spectral,chen2018detecting,li2021neural}. Yet, no definite solution has been found, leading to an ongoing arms-race ~\citep{tan2019bypassing,nguyen2021wanet}.
In contrast to many works in the area of backdoors or poisoning in general, we do neither propose a defense nor an attack. Instead, we study the phenomenon of backdoors in relation to local sensitivity. Along these lines, \cite{frederickson2018attack} show a trade-off in strength of the attack and detectability for general poisoning. Analogously, \cite{cina2021backdoor} show that learning a backdoor requires the model to globally increase complexity if the model's complexity is not high enough. In contrast to both works, we study the local effect that backdoors have on a classifier.
Furthermore,
 \cite{baluta2019quantitative} study backdoor generalization using their formal framework in binary neural networks. They conclude that the trigger is only effective when combined with images from the training distribution. We have instead compared in distribution test data with and without trigger.  
 Finally, \cite{zhu2020gangsweep} investigate the loss surface when generating a trigger and show that a GAN outperforms conventional, gradient based methods. Although gradients are correlated with local sensitivity, the authors focus on the problem of recovering the trigger from clean data, and do not investigate the gradients with respect to the type of data (e.g., with and without trigger).

\paragraph{Randomized Smoothing} Our measure is conceptually related to randomized smoothing, a certified defense against test-time attacks on models introduced by \cite{DBLP:conf/icml/CohenRK19}. Analogous to \mEntr, a smoothed classifier is approximated using Gaussian noise and its class probabilities are computed. However, the authors derive a bound on adversarial robustness to $\ell_2$ perturbations from the probabilities, expressed as a radius in which the prediction of the model does not change. In this area, the classifier is naturally smooth, yielding a strong connection to \mEntr.
Randomized smoothing has been extended beyond Gaussian noise to encompass other metrics such as Wasserstein~\citep{levine2020wasserstein} and specific network architectures, like for example graph neural networks~\citep{jia2020certified}. Extending \mEntr along these lines is thus straight forward.

\paragraph{Measures of Input Sensitivity} A detailed overview about both empirical and theoretical measures of input sensitivity and overfitting is given by \cite{DBLP:conf/iclr/JiangNMKB20}. 
Some of these measures are conceptually close to \mEntr. For example, \cite{forouzesh2020generalization} introduce a measure that is also based on measuring the output variation for a sample perturbed using Gaussian noise. Further, \cite{novak2018sensitivity} propose a measure based on the norm of input/output Jacobian. Both measures, however, are formally connected to overfitting, and conceptually work in a setting where several classifiers are trained on differently perturbed training data.
Finally, \cite{shu2019sensitivity} also propose a perturbation based measure. They empirically show that their measure exhibits differences for training and test data, related to a specific test time perturbations. We, instead, focus on backdoor trigger in our analysis, and do not constrain the perturbation to the manifold of the network. We further aim to understand the differences of input sensitivity for different groups of data, namely with and without active trigger at test time.

\section{Conclusion and Future Work} %0.5 pages
\label{sec:conclusions}
To tackle the lack of understanding in backdoor attacks in deep learning, we studied backdoors  from the perspective of local smoothness of the decision function.
To this end, we proposed a measure, \mName, that quantifies this local smoothness of the decision surface.
Our experiments showed a phenomenon we call backdoor smoothing: the decision surface is smoother around points with an active backdoor trigger pattern than around benign training points.
More concretely, the smoothness around data points with the trigger pattern correlates with the trigger's accuracy.  
We further tested whether backdoors are a necessary condition for backdoor smoothing. To this end, we  confirmed that active backdoor triggers are reliably living in smooth areas of the decision function. 
However, we were also able to craft smoothness-inducing patterns 
that were not previously inserted in the training data. These samples are adversarial examples, and defending them is of independent interest. 

Our findings imply that $\mName$ might be used as a defense against backdoors and also test-time perturbations that induce smoothness. We leave an in-depth evaluation of this defense for future work.
Furthermore, our work is currently limited to the area of computer vision. However, as work in randomized smoothing shows, an extension of $\mName$ to other domains is straight forward.
We also did not investigate how the decision surface changes in terms of smoothness when additional defensive measures are applied. The results from these experiments would for example yield insights about how to combine different mitigations. 
Finally, our work does not encompass the design of an adaptive attacks against backdoor smoothing.
We leave it for future work to answer the question whether such an attack is able to decouple backdoor smoothing and attack success, in particular without affecting clean accuracy.

 \begin{comment}
 \section*{Declaration of Competing Interest}
The authors declare that they have no known competing financial interests or personal relationships that could have appeared to influence the work reported in this paper.

\section*{CredIT Authorship Contribution Statement}
\textbf{Kathrin Grosse}: Conceptualization, Data curation, Formal analysis, Investigation,  Methodology, Software, Validation, Visualization, Roles/Writing - original draft, review and editing. 
\textbf{Taesung Lee}: Conceptualization, Data curation, Formal analysis, Investigation, Methodology, Software, Validation, Visualization, Roles/Writing - original draft, review and editing.
\textbf{Battista Biggio}: Formal analysis, Funding acquisition, Investigation, Methodology, Project administration,  Resources, Supervision, Validation, Visualization, Roles/Writing - original draft, review and editing.
\textbf{Youngja Park}: Conceptualization, Investigation, Methodology, Validation, Visualization, Roles/Writing - original draft, review and editing.
\textbf{Michael Backes}: Funding acquisition, Project administration,  Resources, Supervision.
\textbf{Ian Molloy}: Conceptualization, Funding acquisition, Investigation, Methodology, Project administration, Resources, Supervision, Validation, Visualization, Roles/Writing - original draft, review and editing.
\end{comment}

\section*{Acknowledgements}
Kathrin Grosse was supported by the German Federal Ministry of Education and Research (BMBF) 
through funding for the Center for IT-Security, Privacy and Accountability (CISPA) (FKZ: 16KIS0753). Kathrin Grosse and Battista Biggio are funded by BMK, BMDW, and the Province of Upper Austria in the frame of the COMET Programme managed by FFG in the COMET Module S3AI. Taesung Lee, Youngja Park and Ian Molloy are funded by IBM, Michael Backes by the CISPA Helmholtz Center for Information Security.

\bibliographystyle{model2-names}\biboptions{authoryear}
\bibliography{main.bib}

\end{document}